\documentclass[conference]{IEEEtran}
\IEEEoverridecommandlockouts
\usepackage{cite}
\usepackage{amsmath,amssymb,amsfonts}
\usepackage{algorithmic}
\usepackage{graphicx}
\usepackage{textcomp}
\usepackage{xcolor}
\def\BibTeX{{\rm B\kern-.05em{\sc i\kern-.025em b}\kern-.08em
    T\kern-.1667em\lower.7ex\hbox{E}\kern-.125emX}}

\begin{document}

\title{Agentic Meta-Orchestrator for Multi-task Copilots}


\author{\IEEEauthorblockN{1\textsuperscript{st} Xiaofeng Zhu}\thanks{The open-source version of this work was performed independently following the Xiaofeng’s employment, without corporate affiliation.}
\IEEEauthorblockA{\textit{Microsoft Corporation}\\
Redmond, U.S. \\
Xiaofeng.Zhu@microsoft.com}
\and
\IEEEauthorblockN{2\textsuperscript{nd} Yunshen Zhou}
\IEEEauthorblockA{\textit{Microsoft Corporation} \\
Redmond, U.S. \\
yunshenzhou@microsoft.com}
}
\maketitle

\begin{abstract}
Microsoft Copilot suites serve as the universal entry point for various agents skilled in handling important tasks, ranging from assisting a customer with product purchases to detecting vulnerabilities in corporate programming code. Each agent can be powered by language models, software engineering operations, such as database retrieval, and internal \& external knowledge. The repertoire of a copilot can expand dynamically with new agents. This requires a robust orchestrator that can distribute tasks from user prompts to the right agents.
In this work, we propose an Agentic Meta-orchestrator (AMO) for handling multiple tasks and scalable agents in copilot services, which can provide both natural language and action responses. We will also demonstrate the planning that leverages meta-learning, i.e., a trained decision tree model for deciding the best inference strategy among various agents/models. We showcase the effectiveness of our AMO through two production use cases: Microsoft 365 (M365) E-Commerce Copilot and code compliance copilot. M365 E-Commerce Copilot advertises Microsoft products to external customers to promote sales success. The M365 E-Commerce Copilot provides up-to-date product information and connects to multiple agents, such as relational databases and human customer support. The code compliance copilot scans the internal DevOps code to detect known and new compliance issues in pull requests (PR).

\end{abstract}

\maketitle


\section{Introduction}
\label{Introduction}
Microsoft copilots are designed to handle multiple work tasks by coordinating large language models (LLMs) \cite{dubey2024llama3herdmodels,abdin2024phi3technicalreporthighly,mohamadi2023chatgptagegenerativeai,Dong_2022}, Microsoft 365 productivity apps, and more. LLMs are impressive at handling general tasks, such as drafting documents, question answering, and summarization. Due to data compliance and boundaries, fine-tuning of customer data with web-based LLMs, such as ChatGPT, can be prohibited. Therefore, keeping the knowledge of an LLM up-to-date can be challenging.

\begin{figure}[!b]
\begin{center}
\includegraphics[scale=0.45]{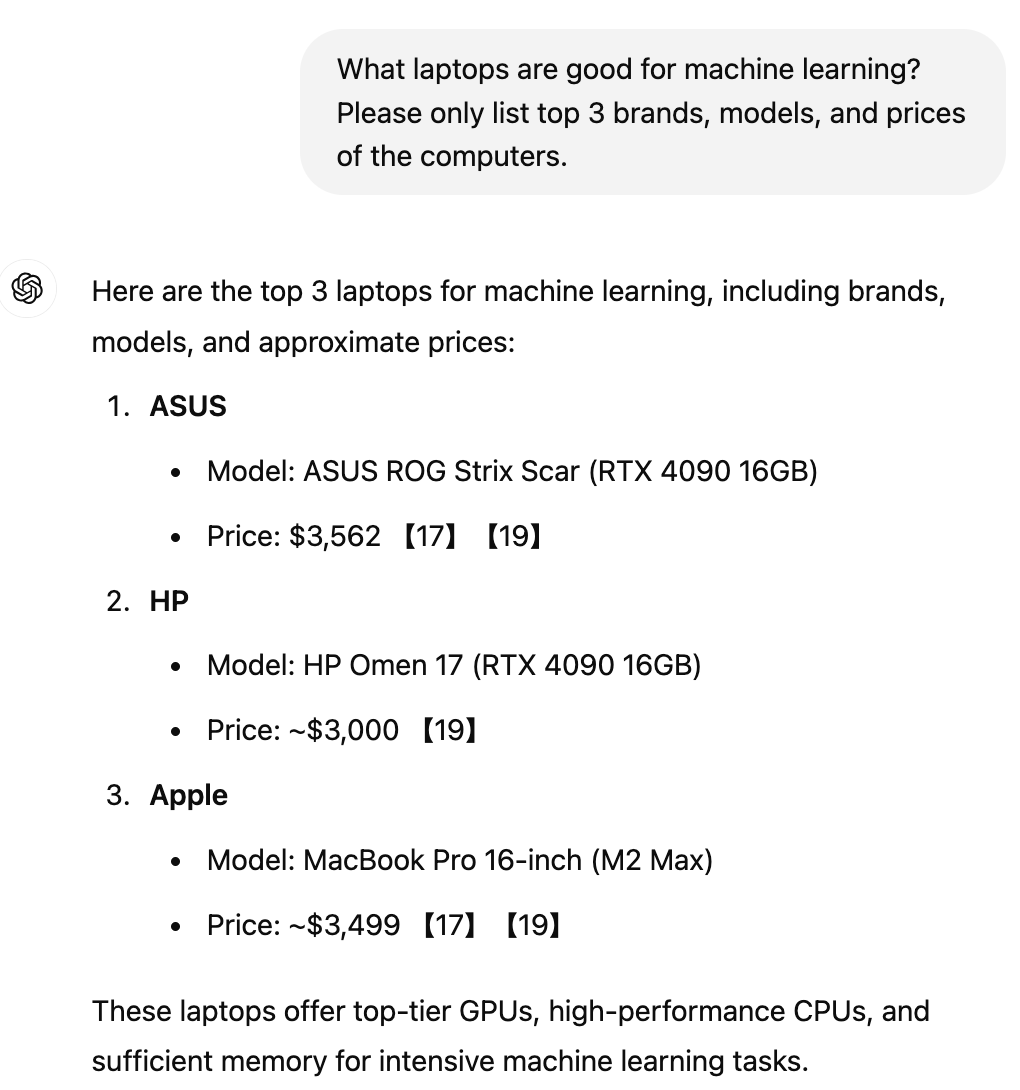}
\caption{Product Recommendation from ChatGPT}
\label{AI-laptops}
{ChatGPT failed to provide detailed or up-to-date information}
\end{center}
\end{figure}

Figure \ref{AI-laptops}
shows a ChatGPT laptop recommendation to a user who plans to train machine learning models. It is easy to find that ChatGPT nicely shows options of ASUS, HP, and Apple models but with outdated information. For example, in 2024 the newest model for an Apple Macbook uses M4 chip instead of M2 shown by ChatGPT, and the prices for ASUS and HP laptops are also different from what are shown online. An E-Commerce Copilot customized for Apple Inc. likely aims to advocate for Apple products instead of ASUS or HP's. In addition, the copilot needs to always fetch the latest information of products, such as pricing, model configurations during promotion seasons. 

\begin{figure*}[pt]
\begin{center}
\includegraphics[scale=0.35]{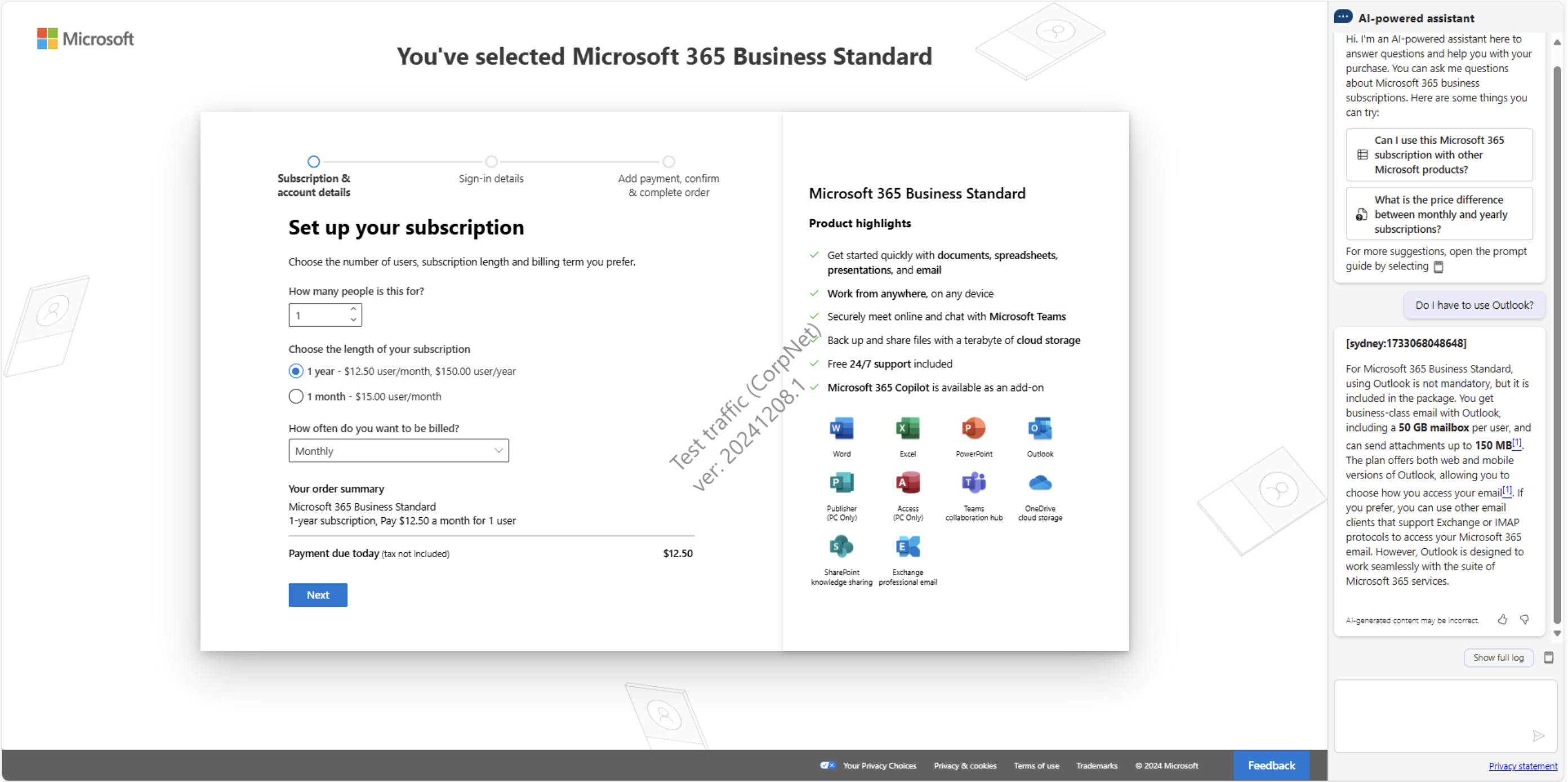}
\caption{Product Information Response from our M365 E-Commerce Copilot}
\label{cart-copilot-outlook}
{Our E-Commerce Copilot provides tailored responses to user prompt ``Do I have to use Outlook?" based on the web page information and shopping status.}
\end{center}
\end{figure*}

\begin{figure*}[pt]
\begin{center}
\includegraphics[scale=0.45]{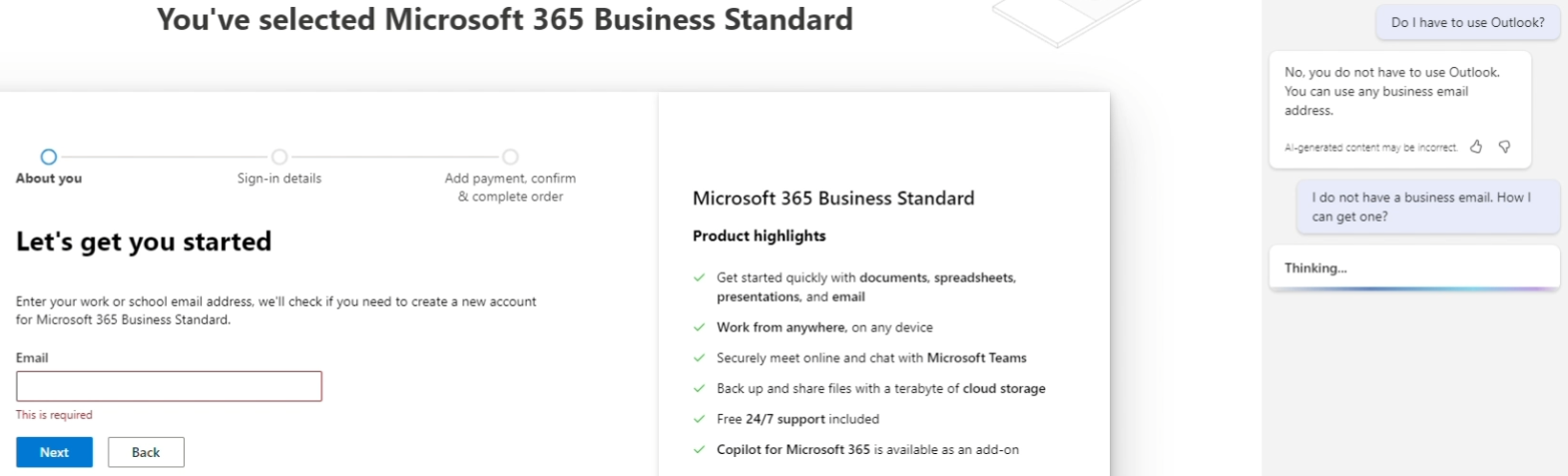}
\caption{Technical Assistance Response from our M365 E-Commerce Copilot}
\label{cart-copilot-outlook-input-field}
{Our E-Commerce Copilot provides tailored responses to user prompt "Do I have to use Outlook?" given that the user is filling in a requested email address.}
\end{center}
\end{figure*}

\begin{figure*}[pt]
\begin{center}
\includegraphics[scale=0.35]{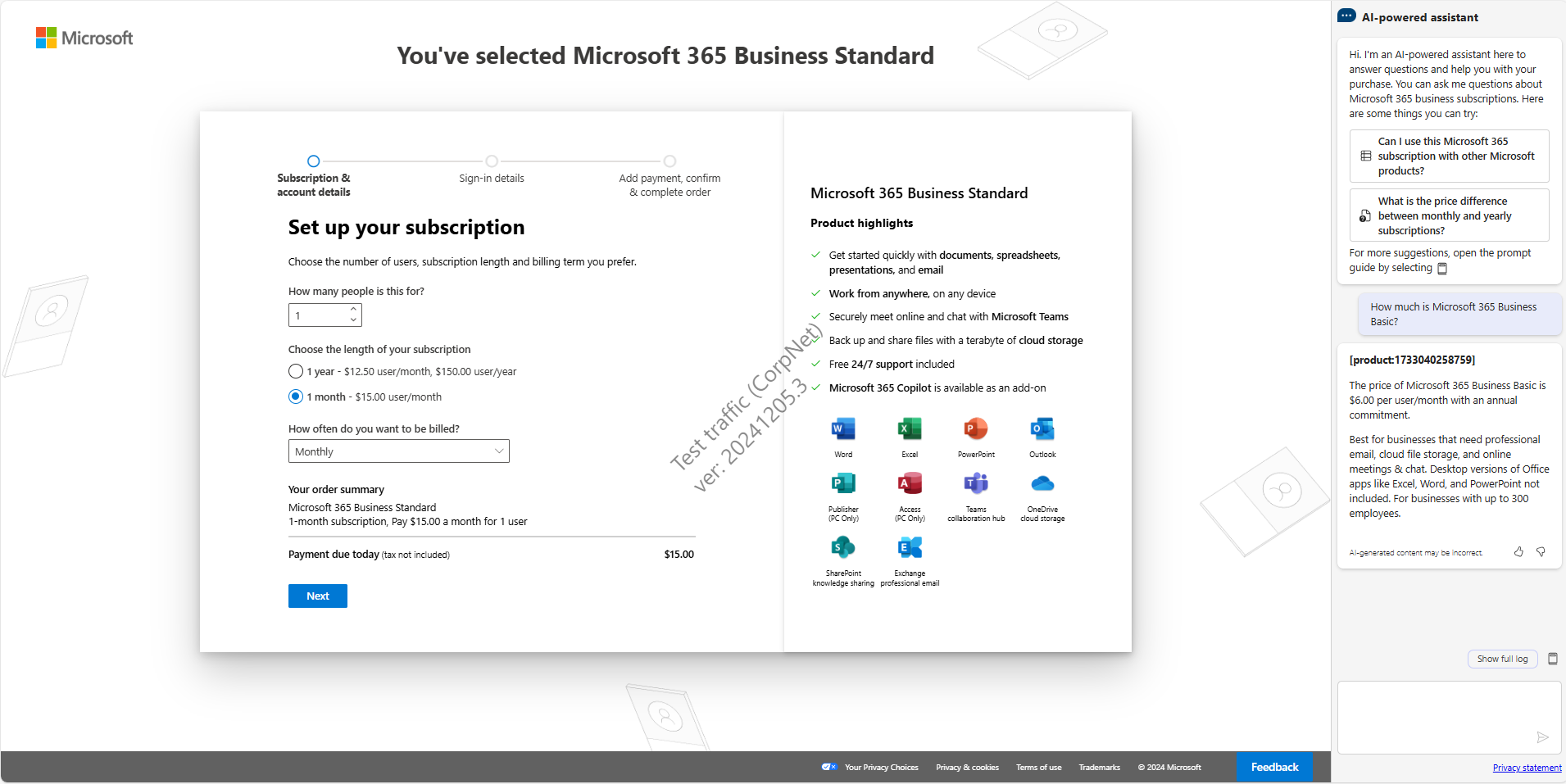}
\caption{Up-to-date Pricing Response from our M365 E-Commerce Copilot}
\label{cart-copilot-demo}
\end{center}
\end{figure*}

\begin{figure*}[pt]
\begin{center}
\includegraphics[scale=0.55]{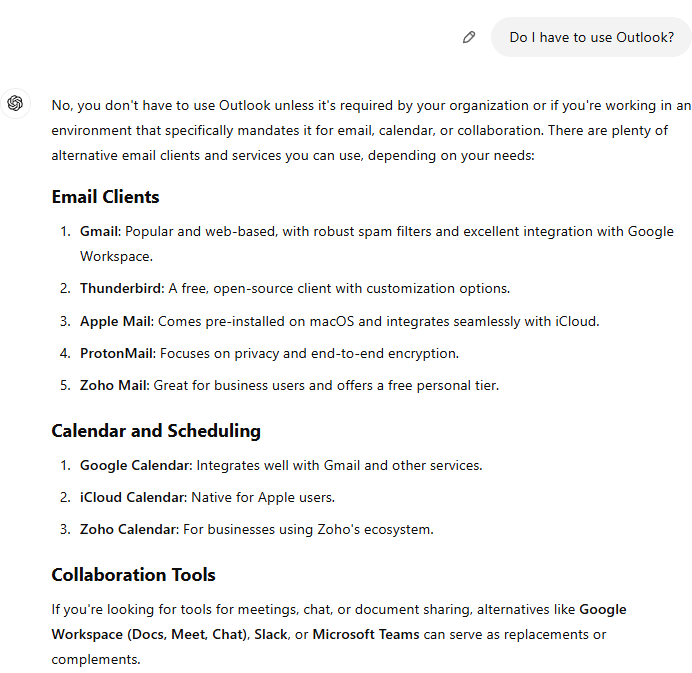}
\caption{General response from ChatGPT}
\label{chatgpt-outlook}
{ChatGPT failed to provide scenario-specific answers to user prompt "Do I have to use Outlook?"}
\end{center}
\end{figure*}

In contrast, our E-Commerce Copilot shown in Figure \ref{cart-copilot-demo} can timely obtain the latest pricing information based on geographic locations (different countries have different prices and currencies for the same product). Moreover, depending on the context or different users, we provide customized responses (for the same question) for different user interactions. Figure \ref{cart-copilot-outlook} provides product information related to Outlook as the customer is reviewing the package details. Figure \ref{cart-copilot-outlook-input-field} demonstrates that our E-comerce copilot understands the context in which the customer is asking if filling the text field on the left side using an Outlook email address is necessary, and it is able to interpret multi-turn messages well. Not surprisingly, Figure \ref{chatgpt-outlook} shows that a standalone ChatGPT is not capable of providing such customization. This issue also occurs in multiple industries, such as instant airline and hotel pricing. 
We relieve the proposed Agentic Meta-orchestrator (AMO) behind the scene in this work. This method not only works on Microsoft products but also can be easily applied to other multi-agent AI systems in various industries.

To provide a customized in-domain assistant for Microsoft customers, language agents can neatly extend the generalizability of LLMs in private domains \cite{10.1145/3627673.3679105,yao2024tau,Wang_2024,zhou2023agentsopensourceframeworkautonomous} and connect with applications, knowledge graphs, databases, other copilots, etc. The explosion of adapting various agents in particular tasks or domains poses challenges for 1) an effective orchestrator that decomposes tasks from user prompts and assigns them to growing agents, 2) efficient deployment of the foundation model and agents, and 3) inference planning among agents, e.g., which agents should be used next, and what messages can be shared across agents.

Regarding the first challenge, we propose a trained multi-level rating learning-to-rank model (with graded relevance) to orchestrate the prompts to the right agents, which is generally taken as a hierarchical text classification problem \cite{zhang2024teleclass,vaswani2023attentionneed,chen2019bert}. The agents can be ``ask for price", ``compare products", ``contact human customer support", etc. as in our M365 E-Commerce Copilot, and can be "encoding character issue", "new character issue" as in the code compliance copilot. An orchestrator can be at the copilot level, e.g., it selects the right plugins \footnote{https://learn.microsoft.com/en-us/microsoft-365-copilot/extensibility/orchestrator} with agents, and it can select the right models based on the business intents of user prompts. As the number of agents continues to grow, the use of a text classification model \cite{liu2019robertarobustlyoptimizedbert,devlin-etal-2019-bert} can have difficulties in scalability, and the cosine similarity method can struggle when the descriptions of agents have overlaps or when user prompts are ambiguous. 

Regarding the second challenge, we propose a memory-efficient inference framework, LoRA\cite{hu2021loralowrankadaptationlarge} arms, where each arm handles one task. Hosting multiple LLMs requires substantial memory\cite{vyas2024autonomousindustrialcontrolusing,zhang2024chainbuddyaiagentgenerating}. 
Inspired by the LoRA framework, we share the base LLM model among tasks and train LoRA weights for individual tasks. 

Regarding the third challenge, we propose a novel meta-learning\cite{lee-etal-2022-meta} decision tree model that trains the best inference model combinations and orderings for different prompts. The well-known successful framework, retrieval-augmented generation (RAG), is a simplified two-step inference planning based on human heuristics that can be taken as a web search agent and LLM question-answering agent. When a complex copilot service is composed of multiple agents and combinations of skills/models, how we effectively choose an inference strategy for individual user prompts requires a meta-learning approach. Current planning strategies, such as AutoGen\cite{dibia2024autogen,porsdam2024autogen,wuautogen,zhu2023autogen,wu2023autogen}, CrewAI, and TaskWeaver\cite{barbarroxa2024benchmarking}, rely on agentic workflows and task-specific prompts, which are generally based on cognitive architectures\cite{sumers2023cognitive,ghosal2025doesthinkinghelpunderstanding,deng2025simura,kang2024self}.

We summarize our contributions as follows.\\
1) We convert the hierarchical text classification to a multi-level relevance rating semantic learning-to-rank task taking natural language descriptions of labels instead of using a softmax, which is challenging to scale up when new agents/classes are added in.\\
2) We can locate top-k based on individual user prompts by adjusting the ranking position of the "separator" class or candidate document.\\
3) We productize LoRA arms for efficient inference of multiple tasks. Each arm comes from an independent LoRA fine-tuning, which contributes to the scalability and flexibility of complex AI systems, such as copilots. \\
4) We present a meta-learning model for identifying the optimal inference planning given various combinations of task-specific agents/models. Our independently developed, non-corporate code and data used in Section \ref{Extension to Hierarchical Text Classification Results} are made publicly available \footnote{https://github.com/XiaofengZhu/AMO} for research purposes, without implementing any patented methods or systems.







\begin{figure*}[pt]
\begin{center}
\includegraphics[scale=0.55]{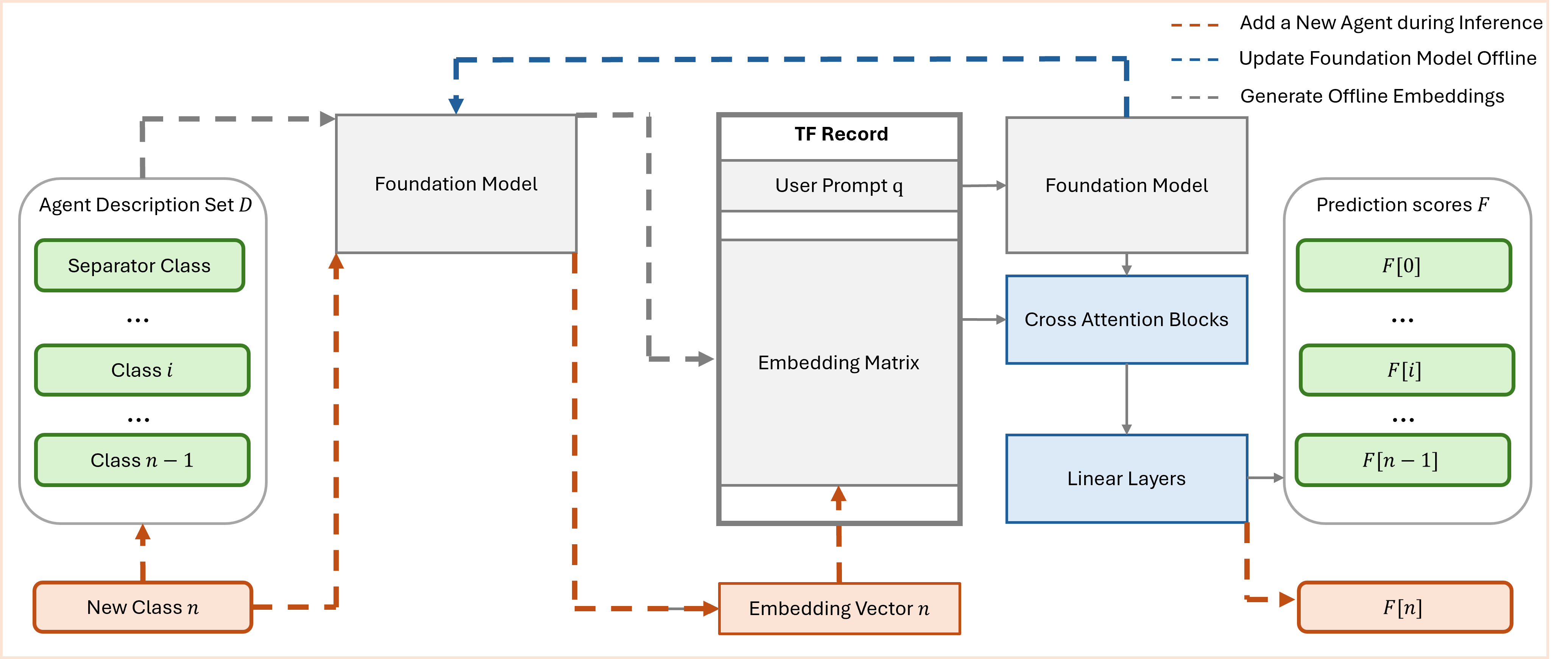}
\caption{Agentic Orchestrator}
\label{sRankOrchestrator}
{Hierarchical text classification is a multi-level rating learning-to-rank task} 
\end{center}
\end{figure*}

\section{Related Work}
\label{Related Work}

\subsection{Task Distribution and Semantic Learning-to-rank}
\label{Task Distribution and Semantic Learning-to-rank}

LLMs have been attempted in few-shot text classification tasks\cite{li-etal-2024-pdameta,zhang-etal-2022-prompt-based} and few-shot learning-to-rank models\cite{qin-etal-2024-large,jiang-etal-2023-llm,guo2022rankdnnlearningrankfewshot,sinhababu2024fewshotpromptingpairwiseranking}

Previous studies have shown that learning-to-rank can be used explicitly and implicitly in various language tasks \cite{10.1145/3543873.3584621,karpukhin-etal-2020-dense}. However, they were limited to binary ratings. We apply semantic learning to rank to multi-level rating scenarios. We utilize the natural description of classification labels as the learning-to-rank ``candidate documents", different from the $0$ to $n$ classes using one-hot encoding in softmax.

\subsection{Multi-task Learning}
Multi-task learners generally share a common module \cite{zhang-etal-2023-survey, chen2019bert}. 
Previous studies, such as MT-DNN\cite{bhattacharjee2022multendtoendmultitasklearning} and MulT\cite{liu2019multi} have focused on training multiple tasks at the same time by sampling one task at a time using a neural network with shared layers and task-specific layers. 
This method can suffer performance drops in subsets of tasks, and it is not easy to efficiently update models based on individual task requirements. The authors of \cite{stickland2019bert} proposed projected attention layers (PALs) in parallel to each layer for BERT \cite{devlin-etal-2019-bert}, which learns task-specific information in the additional parameters. The idea is similar to LoRA, which uses low-rank matrices to reduce the number of parameters for attention in task-specific layers.  However, when we update the training data for one task, the performance of other tasks also changes, which is not good for updating the models, especially considering in many cases in industry we want to update the model for one task without affecting the performance for other tasks.
We show that we achieve impressive individual and end-to-end task performance leveraging LoRA fine-tuning and inference memory optimization in Sections \ref{LoRA Arms} - \ref{Meta Planning} and \ref{Experimental Results}.

\subsection{Agentic Workflow}
\label{Agentic Workflow}

Unlike chain of thoughts\cite{wei2023chainofthoughtpromptingelicitsreasoning}, tree of thoughts \cite{yao2024tree}, or tree search \cite{light2024strategistlearningstrategicskills,yu2024exactteachingaiagents} that address LLM inference planning, our meta-learning planning strategy is a trained model of models/agents.

A well-known cognitive heuristics RAG \cite{edge2024localglobalgraphrag} plans a knowledge retrieval then LLM generation. Although great reinforcement learning approaches, for example, React \cite{yao2022react} and Reflection \cite{shinn2023reflexionlanguageagentsverbal}, and Self-critic \cite{saunders2022selfcritiquingmodelsassistinghuman} have been attempted for complex systems, how we quickly update the planning when new agents are added remains an open problem. Our proposed AMO novelly solves this problem of continually adding new agents.

Another good cognitive heuristic is ensemble modeling. Each record is processed by the same models using bagging, stacking, etc. Our AMO, in contrast, teaches the methodology of training a model \cite{ha-etal-2023-meta,10.5555/3600270.3602999,chen-etal-2022-meta,zhang-etal-2022-prompt-based} and is Conversable; agents can share information with each other.

\section{Proposed Agentic Meta-orchestrator}
\subsection{Agentic Orchestrator of Top-k Selections}
\label{Agentic Orchestrator of Top-k Selections}

Let us assume that there are three task-specific agents available for a copilot to orchestrate user prompts to. Training a multi-class text classification model is an intuitive approach. However, we deal with a growing number of agents that are added to copilots on a regular basis. For example, copilot partners can request for on-boarding their new agents of handling tasks in private domains, e.g., creating a new PowerPoint deck, checking order status, adding a user to a purchased subscription, etc.
It is not user-friendly to have newly added agents wait until a new fine-tuned model that includes this class is available to route user prompts to them. In addition, it is uneasy to choose the cutoff top-k value if we search based on key words or sentence embeddings.

Therefore, we design an orchestrator that can seamlessly take new agents on the basis of their agent/class labels. Figure \ref{sRankOrchestrator} demonstrates that we convert agent/class descriptions to semantic embeddings offline and use them to train a listwise multi-level learning-to-rank model \cite{10.1145/1963405.1963459,chen-etal-2020-hierarchical,10.1145/3336191.3371814}. The model targets higher relatedness prediction scores for agents that are ranked higher than others. In particular, when there is only one correct item in a ranking list or classification group, such as `[0, 1, 0, 0]', the gradients of ListMLE-style loss functions and Softmax are equivalent. However, the former is more applicable to multi-level ratings or hierarchical classes. The user prompt and the agents use the same embedding model, which is generic. We were able to achieve stable performance using BERT and Sentence Transformers as the foundation models, even more stable than using softmax classification trained for all classes. 

Each batch contains the trainable embedding of $q$ a user prompt, fixed embeddings of $D^q = \{d\}$ the set of candidate agent descriptions associated with $q$, which can differ in batches, and $R(D^q) = \{r(d)\}$ the relatedness rating for $D^q$, ranging over all real numbers (showing that the previous study in \cite{10.1145/3336191.3371814} can be generalized).
We maximize the likelihood of selecting a document or class $d$ from $c$ = $d \cup \tilde{s}$:

\begin{equation} \label{P_t(d)}
P(d) = \frac{exp(f(d))}{exp(f(d)) + \sum_{d' \in \tilde{s}} exp(f(d'))}.
\end{equation}

\noindent The rating of every document in $\tilde{s}$ is less than $r(d)$\cite{10.1145/3336191.3371814}. We finalize our general learning to rank loss function uRank loss by adding a weight factor of $2^{r(d)} - 1$. We list and answer the following core questions accordingly.

1) \textbf{What are class hierarchies?}\\
The semantics of real-world agents carry hierarchies. For example, for an E-Commerce Copilot, it detects the business intents of user prompts and only continues processing intents that are related to the company's products, then routes the prompts to the desire categories varying from Office products, Azure products, Xbox, etc.
In addition, high-level copilots generally have limited data access to private domains. 

2) \textbf{What are the limitations of the classical softmax multi-class text classification?}\\
The labels of a three-class classification model can be represented as $[0,1,0]$. When the fourth class comes in, we need to extend the labels to e.g., $[0,0, 1,0]$. Therefore, we generally need to train a new multi-class text classification model when a new class appears. 
Moreover, the $0-1$ labeling system assumes that all classes are independent, which ignores the semantics of the class labels and the connections between the labels and the training texts. Such limitations impact the scalability of copilots as they continually onboard new agents.

3) \textbf{How can we use learning-to-rank to orchestrate a growing number of agents?}

As we have learned about the limitation of the sofmax text classification, we utilize agent/class descriptions, such as ``ask for price," as the ``candidates" and train a learning-to-rank model to select top-k agents given a user prompt. All user prompts may share the same or different number of agent/class descriptions. A learning-to-rank model can support this flexibility. 

4) \textbf{How can we define a multi-level rating?}\\
Agent ratings are defined by hierarchical levels and relatedness to user prompts. For instance, for a user prompt ``How much is M365 Business Standard annual subscription?" the ratings of agents ``Ask for Price", ``Microsoft products", ``Separator Class", ``Compare Products", ``Non-Microsoft Products" are $\{2, 1,0,-1,-2\}$, where the higher the positive rating the agent is more related to the user prompt, the lower the negative rating the agent is less related to the user prompt. ``Separator Class" uses ``other Microsoft product matters" as the agent description.
For a different user prompt "What apps are included in M365 Business Standard?" the ratings of agents ``Separator Class", ``Microsoft Products", ``Ask for Price", ``Compare Products", ``Non-Microsoft Products" are $\{2, 1, 0, 0, -1\}$ based on their natural semantic hierarchies.

5) \textbf{How can we choose the top-k agents/agents based on user prompts?}\\
We add a ``Separator Class" that indicates a class that is not defined in current categories. The learning-to-rank model maps the Separator Class to the top-k position, i.e., the placeholder Separator Class itself and agents/classes that are ranked below this Separator Class are not relevant to the user prompts. 

6) \textbf{Why do we choose multi-level rating learning-to-rank over text generation?}\\
Recent studies have good experiments in LLM-based few-shot text classification. Though LLMs consider semantics of class labels, LLM generations can be more creative than we need, i.e., they do not guarantee that the outputs contain the defined class labels. We have compared with this approach and detailed the differences in Table \ref{Copilot Results}. Moreover, due to the token limit for LLM we may not feed enough few-shot examples into the prompt, leading to poor performance.

\begin{figure*}[ht]
\begin{center}
\includegraphics[scale=0.55]{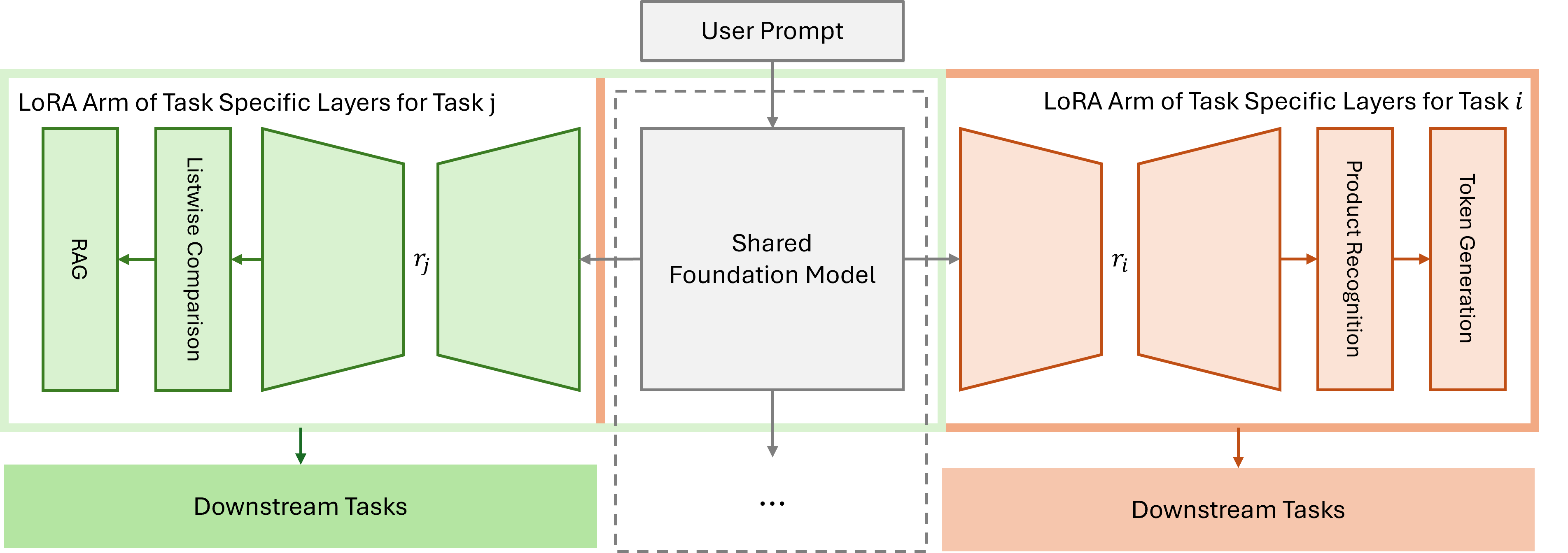}
\caption{LoRA Arms of Handling Multiple Tasks}
\label{LoRA-Arms}
{LoRA-Arms are trained independently during fine-tuning and can be used simultaneously during inference.}
\end{center}
\end{figure*}

\begin{figure*}[pt]
\begin{center}
\includegraphics[scale=0.55]{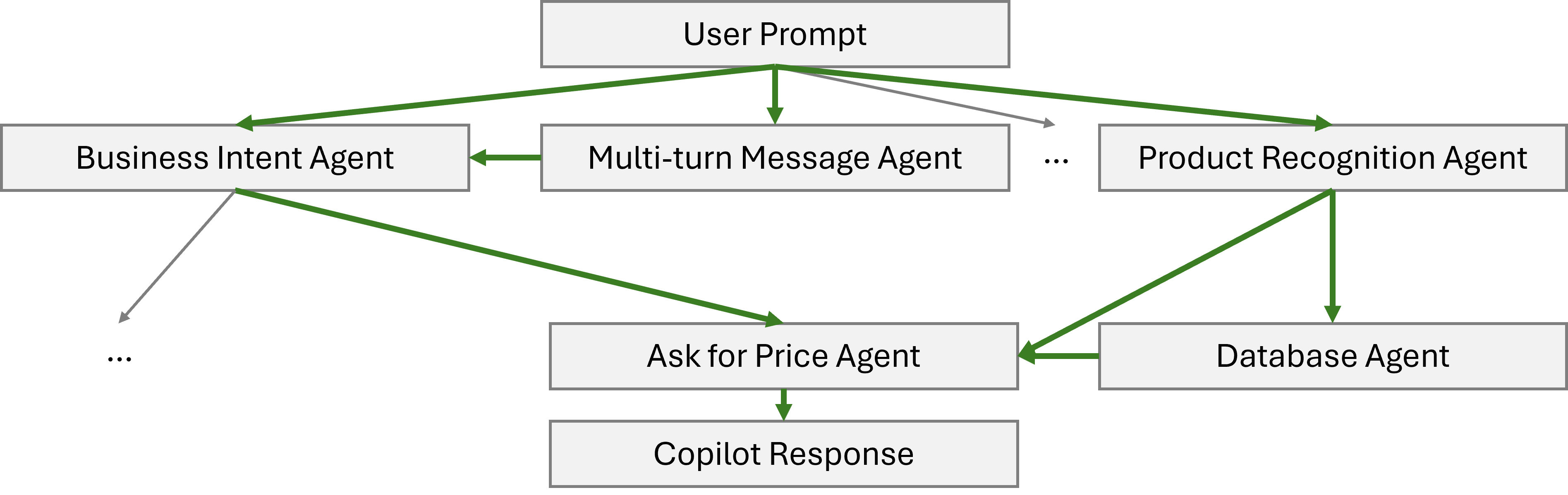}
\caption{A Meta-learning Decision Tree Model, in which Each Node is a Language Agent or Model}
\label{Meta-Learner}
{The green path shows the inference planning for ``How much is M365 Business Standard Annual Subscription?" utilizing ``Business Intent", ``Multi-turn Message", ``Product Recognition", ``Database" and ``Ask for Price" agents}
\end{center}
\end{figure*}

\subsection{LoRA Arms}
\label{LoRA Arms}

One major issue of leveraging multiple agents/models at the inference stage is memory consumption. Hosting a service with multiple LLMs fine-tuned from the same foundation model causes memory usage waste.

Slightly different from LoRA memory optimization that subtracts the old LoRA weights and then adds the LoRA weights of the new task in the study by Hu et al. \cite{hu2021loralowrankadaptationlarge}, we support multiple LoRA arms during inference at the same time, as shown in Figure \ref{LoRA-Arms}. They share the same LLM memory as long as the compute resources allow.

The user prompts are passed to agents, either sequentially or asynchronously in parallel. When the same foundational models make sequential LLM calls or perform inference across multiple LLMs fine-tuned from the same base model, it can lead to significant waste of compute resources—particularly memory consumption—which directly impacts the scalability of the overall systems or copilots.

Core tasks in our copilots include (implicit) product entity recognition, reflection of consuming long-term and short-term memory/context in multi-turn messages of a chat conversation. Each was trained and predicted using a LoRA arm. This is essential for supporting multiple application programming interface (API) services that share the foundation models in memory, which are used in different downstream workloads.

\subsection{Meta Planning}
\label{Meta Planning}
Instead of responding solely to cognitive heuristics, we have a meta-learning decision (tree) model to decide on the best inference planning paths for user prompts. Training inputs are user prompts, their end-to-end copilot responses, and various models for agent-specific tasks. The inference planning includes the combination and ordering of agents, e.g. only using Phi-3.5\cite{abdin2024phi3technicalreporthighly} or going through product recognition agent, database agents, etc. Slightly different from classical decision trees, we allow visiting the same nodes and paths several times by enabling retries during the inference stage. Then default paths, e.g., human customer service pop-up windows, and pre-calibrated messages, will be triggered.

Figure \ref{Meta-Learner} shows the meta-learning decision (tree) model for our copilots. Each node is an AI language agent, and the orchestrator assigns the current user prompt to complete certain tasks. The planning decisions of the Meta-learner need to be actionable, e.g., proceeding with an exact agent with specific settings, such as English and U.S. The node can be whether a Microsoft product entity, such as Teams Essential, Microsoft 365 Business Premium, is recognized by the product recognition agent, and how many product entities are detected. AMO includes BERT, Phi-3.5 and their task-specific LoRA arms. The planning paths are deterministic during the interactions with customers after we learn the best actions from the training data. Therefore, reinforcement learning\cite{deng2025simura,jha2024neural,hou2504thinkprune} was not the focus of our current study.

\section{Production Use Case Experiments}
We zoom in on the two production copilot services: M365 E-Commerce Copilot and Code Compliance Copilot. We have conducted benchmarking experiments for the agent orchestration task for the two copilots, and the multi-task handling for the M365 E-Commerce Copilot. In the end, we compare with AutoGen, a prompt workflow engine with different agents designed for our tasks. The gains of AMO over standalone LLMs—Fine-tuned Phi-3.5 and ChatGPT-4o—reflect the impact of the proposed agentic orchestrator and planning. The improvement of AMO over AutoGen demonstrates the effectiveness of AMO’s meta planning compared to LLM planning and context understanding in multi-turn conversations.

\subsection{M365 E-Commerce Copilot}
\label{M365 E-Commerce Copilot}
M365 E-Commerce Copilot helps Microsoft customers choose the right products and technical configurations. The dataset contains around $30k$ user prompts with labeled agent classes, such as ``Ask for Price", ``Non-Microsoft Products", ``Contact Human Support Agents", and labeled product entities, e.g., ``Teams Essential", ``Excel."

\subsection{Code Compliance Copilot}
\label{Code Compliance Copilot}
Code Compliance Copilot scans every PR of internal corporate M365 product code DevOps to make sure that the code changes are compliant, particularly related to Chinese character GB compliance issues. In addition to discovered GB issues, we hope to continue monitoring and identifying new undiscovered GB issues. Therefore, we have hierarchical classes, e.g., GB issue, non-GB issue, GB display issue, GB encoding issue, etc.

\begin{figure*}[pt]
\begin{center}
\includegraphics[scale=0.55]{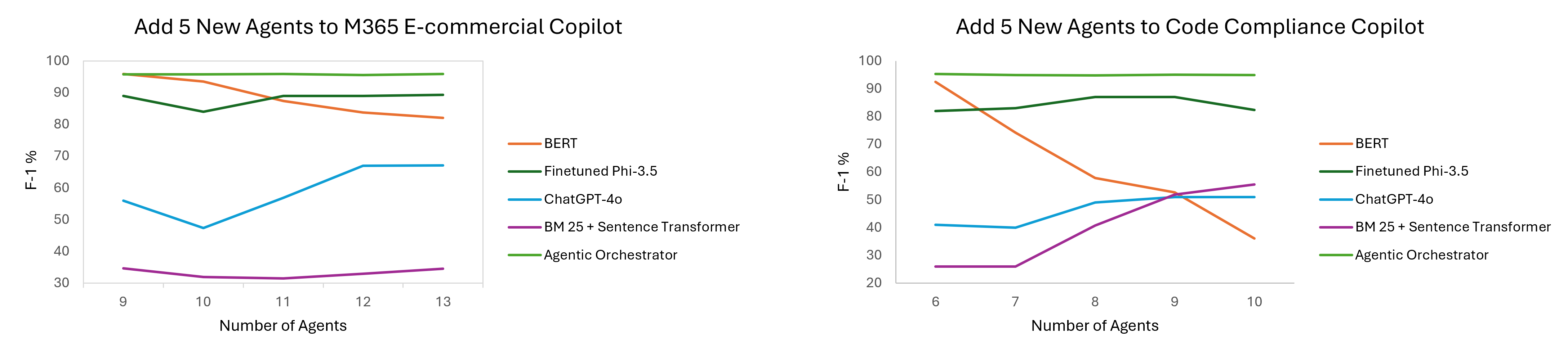}
\caption{Classification F-1 with Growing Agents}
\label{growing-agents}
{Our agentic orchestrator remains stable, while BERT performance drops rapidly}
\end{center}
\end{figure*}

\begin{table*}[pt]
\centering
\scalebox{0.95}{
\begin{tabular}{l|r|r|r|r}
\multicolumn{5}{c}{M365 E-Commerce Copilot Dataset} \\ \hline
Models 	& ROUGE-L 	& BERTScore  & Agent Orchestration Classification F-1 & Product Recognition F-1 \\ \hline
AMO    &\textbf{9.67} &\textbf{5.35} &\textbf{15.93}  &\textbf{26.69}    \\ \hline
Fine-tuned Phi-3.5  &2.41  &-2.33 &8.34  &8.05   \\ \hline
ChatGPT-4o  &-24.91  &-35.89 &-13.67  &-3.95   \\ \hline
AutoGen  &-3.00  &-4.51 &-2.82  &-5.11\\ \hline
\end{tabular}
}
\scalebox{0.95}{
\begin{tabular}{l|r|r|r}
\multicolumn{4}{c}{Code Compliance Copilot Dataset} \\ \hline
Models 	& ROUGE-L 	& BERTScore  & Agent Orchestration Classification F-1 	 \\ \hline
AMO    &\textbf{26.13} &\textbf{30.60} &\textbf{31.02}   \\ \hline
Fine-tuned Phi-3.5  &16.67  &11.52 &12.33   \\ \hline
ChatGPT-4o  &-28.80  &-20.45 &1.92   \\ \hline
AutoGen  &-8.94  &-5.58 &10.55  \\ \hline
\end{tabular}
}
\caption{Performance Gains(\%) of End-to-end M365 E-commerce and Code Compliance Copilots}
\label{Copilot Results}
\end{table*}

\subsection{Experimental Results}
\label{Experimental Results}
We conduct experiments of classification tasks using our proposed agentic orchestrator, BERT, Fine-tuned Phi-3.5, ChatGPT-4o, BM 25 + Sentence Transformer\cite{reimers2019sentencebertsentenceembeddingsusing} for a growing number of agents in Table \ref{growing-agents}. After we train agentic orchestrator, BERT, Phi-3.5 models using 9 out of 13 classes for the M365 E-Commerce Copilot data set and 5 out of 10 classes for the code compliance copilot data set, we inference them on the test data adding 5 new agents one by one. Our agentic orchestrator performs better than Phi-3.5, ChatGPT-4o models indicate that it is more reasonable to treat multi-label or hierarchical text classification as a multi-level rating learning-to-rank task than a generation task.

We show the superior end-to-end performance of ROUGE-L \cite{lin-2004-rouge}, BERTScore\cite{zhang2020bertscoreevaluatingtextgeneration}, and the F-1 measures of multiple tasks using our proposed AMO, Fine-tuned Phi-3.5, ChatGPT-4o, and AutoGen against the baseline of our corporate GPT-3.5 Turbo model in Table \ref{Copilot Results}.

\begin{table*}[pt]
\centering
\begin{tabular}{l|r|r|r}
\multicolumn{4}{c}{Amazon Product Review Dataset} \\ \hline
Models 	& Level-1 Accuracy 	& Level-2 Accuracy  & Level-3 Accuracy 	 \\ \hline
Extension for Agentic Orchestrator   &\textbf{0.96} &\textbf{0.85} &\textbf{0.83}   \\ \hline
Supervised Text Classifiers  &0.95  &0.81 &0.72  \\ \hline
ChatGPT-4  &0.90  &0.75  &0.66  \\ \hline
Phi-4  &0.87  &0.71 &0.59   \\ \hline
\end{tabular}
\caption{Hierarchical Text Classification Accuracy}
\label{Hierarchical Text Classification Results}
\end{table*}

\section{Extension to Hierarchical Text Classification Results}
\label{Extension to Hierarchical Text Classification Results}
To demonstrate why we chose uRank loss as the objective function instead of classification loss functions in our proposed agentic orchestrator, we combined uRank with an MPNet Sentence Transformer \cite{song2020mpnet} and applied this combination to a hierarchical text classification \cite{aly2019hierarchical} data set\footnote{https://www.kaggle.com/datasets/kashnitsky/hierarchical-text-classification}. This Amazon product review dataset contains 40k training records and 10k validation records. There are structured classes of three-levels: 6 ``level 1" classes, 64 ``level 2" classes, and 464 ``level 3" classes, which inherit semantic hierarchies of relevance ratings. In table \ref{Hierarchical Text Classification Results}, we report the superior results of our experimental setup compared to popular supervised text classifiers using transformers \cite{meng2019weakly}, sequence generation few-shot learners\footnote{https://github.com/microsoft/autogen} using Phi-4 model\footnote{https://huggingface.co/microsoft/phi-4} and ChatGPT 4. For sequence generation of few-shots learners, we use the RAG method, which retrieves top-20 related records in training data using embeddings generated by the same Sentence Transformer for fair comparison. While text classifiers require three or more separate models for the three levels, and the sequence generation few-shot learners require multiple LLM calls by design, the levels of classes contribute to the data argumentation of our agentic orchestrator extension. It was interesting to find that, for text classification tasks, the uRank objective function and even the classical softmax-style text classifiers generally outperformed few-shot approaches using LLMs.

\section{Limitations}
Incorporation with online learning leveraging reinforcement learning is an interesting area that we did not focus on for the current production release phases. In particular, we hope that our copilots can take deterministic and consistent actions per user prompts and interactions. Additional signals, such as user interaction, can be leveraged for multi-modal inputs.

\section{Conclusion}
\label{Conclusion}
We have proposed AMO that can orchestrate user prompts to the best agents and can resolve the uncertain top k selections for production autonomy. We have independent LoRA fine-tuning processes for various tasks, and the trained LoRA weights are used for those tasks while sharing the orignal base model during the inference stage. Finally, we demonstrate the efficiency of our meta-learning planner, selecting different inference strategies based on user prompts. We have shown the method is effective in Microsoft Copilot cases such as M365 E-Commerce Copilot and code compliance copilot. This architecture can also be extended to any multi-agent AI systems in various industries such as travel, online merchant, etc. We plan to explore reinforcement learning fine-tuning and choosing different arms \cite{cruz2023reinforcementlearningfinetuninglanguage,10.1145/3546790.3546804}, reasoning and actions among collaborative intelligence\cite{10.5555/3635637.3663215,zhang2024raftadaptinglanguagemodel}, federate learning at the copilot level \cite{10.1145/3664650}.


\bibliographystyle{IEEEtran}
\bibliography{ref}


\end{document}